\documentclass{amia}

\usepackage[export]{adjustbox}
\usepackage{algorithm}
\usepackage[noend]{algpseudocode}
\usepackage{amsmath}
\usepackage{amssymb}
\usepackage{array}
\usepackage{bm}
\usepackage{booktabs}
\usepackage{caption}
\usepackage{subcaption}
\usepackage{etoolbox}
\usepackage{float}
\usepackage{graphicx}
\usepackage{lineno}
\usepackage{listings}
\usepackage{multirow}
\usepackage[normalem]{ulem}
\usepackage{rotating}
\usepackage[table]{xcolor}
\usepackage[labelfont=bf]{caption}
\usepackage{threeparttable}
\usepackage{tabularx}

\usepackage{tikz}

\usetikzlibrary{arrows.meta,positioning,decorations.pathreplacing,calc,shapes.geometric,shapes.symbols,shapes.multipart,shapes.arrows,shapes.callouts,shapes.misc, fit}

\usepackage{mwe}

\newcounter{paranum}

\begin{document}

\title{Clinical Term Extraction using Open-Source Small Language Models}

\author{
Noah Marchal, PhD, MFA$^1,^2$,
William E. Janes, OTD, MSCI, OTR/L$^3$, 
Mihail Popescu, PhD$^1,^2$,
Xing Song, PhD$^1,^2$
}

\institutes{
$^1$ Informatics and Data Science Institute;
$^2$ Biomedical Informatics, Biostatistics and Medical Epidemiology Department; 
$^3$ Department of Occupational Therapy; 
University of Missouri, Columbia, Missouri, USA
}

\maketitle

\section*{Abstract}
{\setlength{\parindent}{0pt}\noindent\itshape
Clinical information for amyotrophic lateral sclerosis (ALS) care documented in unstructured clinical notes limits downstream analysis without extraction into structured formats. 
Open-source small language models with few-shot prompting for detecting the presence of ALS-relevant clinical terms in patient documentation were evaluated without task-specific training data. 
The detection task targeted 17 categories spanning functional scores, respiratory measures, medications, and related clinical and non-clinical attributes. 
Clinical note content was normalized from JSON-encoded discharge summaries and processed with a prompt template having structured JSON outputs. 
We compared 26 open-source models using aggregate, label-level, and manual-validation multilabel classification metrics. 
Manual validation showed that a regex rule baseline had higher overall micro-F1 and lower Hamming loss than any single SLM or TF-IDF baseline, while Qwen3-4B-Instruct-2507 was the highest-performing SLM by micro-F1.
Model rankings varied by metric and label category, with the TF-IDF baseline showing high recall but low precision, some SLMs showing higher precision but lower recall, and Hammer2.1-7b showing strong performance for ALSFRS-R subscore detection.
These findings support targeted hybrid extraction workflows rather than replacement of existing rule-based methods.
\par}

\section*{Introduction}

Electronic health records (EHR) store documentation in provider narrative notes as unstructured or semi-structured text. 
Clinical notes describe patient symptoms, document physical examinations, record medication changes, and may track disease progression through the inclusion of functional assessment scores. 
Narrative documentation is meant to capture clinician observations not represented in structured fields. 
However, the unstructured nature of clinical notes poses a barrier to their secondary use at scale, as relevant information needs to be extracted and transformed into discrete values for downstream computational tasks \citep{Yan2025}. 

Term extraction is particularly pertinent for analyzing data for patients with neurological conditions such as amyotrophic lateral sclerosis (ALS) \citep{Goyal2020}. 
ALS care relies on longitudinal tracking of functional decline through instruments like the ALS Functional Rating Scale-Revised (ALSFRS-R), respiratory measurements such as forced vital capacity, manual muscle strength testing, and histories of symptoms and changes over time \citep{Erb2024}.
Despite efforts such as the CReATe Consortium's ALS Toolkit to promote the use of structured forms to capture all relevant clinical observations for ALS, narrative documentation remains the predominant approach for recording ALS disease parameters \citep{Granit2022}.
Narrative notes have varying formats, terminology, and completeness rates across providers and clinics. 
To extract at scale, methods are needed that can account for linguistic variation, incompleteness, and semantic nuances without manual chart review \citep{Lee2020}.

Current approaches to clinical information extraction make use of natural language processing or supervised machine learning models trained on annotations \citep{Fu2020}. 
Rule-based systems use pattern matching when text is pre-formatted with clearly defined labels but do not easily handle cases with high term variability and those with contextual dependencies. 
Supervised models require significant effort to create annotated labels for training, and the labels may not generalize well across sites with different documentation practices \citep{Magoc2023}. 
Both paradigms require domain expertise for engineering the labels and rules needed for capturing the full intent of the clinician when documenting encounters \citep{Fu2020}.

Small language models (SLMs) present an alternative solution, as they infer outcomes on multiple text formats with few examples \citep{Raffel2023}. 
SLMs encode linguistic terms and semantic relationships using pre-training with large, diverse text sources.
These models are ideal as they don't require task-specific training data or manual labels or extraction rules \citep{Washington2025}.
In contrast to large language models (LLMs), which are often proprietary and require computationally and memory-intensive infrastructure for deployment, SLMs offer a lightweight and accessible alternative. Their smaller parameter size enables efficient local or edge deployment, lower operational costs, and greater transparency, making them particularly useful for privacy-sensitive clinical environments and resource-constrained research settings.

Open-source small language models were investigated for detecting clinical terms relevant to ALS patient care from clinical documentation and extraction. 
As an initial analysis stage, models were evaluated for term presence detection without value extraction. 
By determining whether specific terms or types exist in the documentation, the method could be used to evaluate data completeness, differing documentation practices, and the feasibility and accuracy of subsequent term extraction processes. 
The binary classification task reduces prompt length and allows for pre-structuring outputs in place of extracting exact numerical values or ad hoc structuring of clinical measurements.

Because clinical label categories differ in documentation quality and structure, performance is expected to vary across model families based on their prior training examples.
As such, we compared diverse open-source models and considered whether their strengths might be complementary.
We hypothesized that standardized, consistently documented elements would be easier to detect than sparse or narrative-specific terms. 
The findings may help establish baselines for open-source language model capabilities on clinical term extraction tasks, and also help identify models suited for particular clinical domains.
The results could inform resource decisions for organizations implementing automated clinical information extraction pipelines, as well as future development of hybrid systems combining SLM extraction with rule-based validation.


\section*{Methods}

Patient discharge summaries were represented in two JSON schemas: a keyed section-object format and a DOM-like tree format. 
We extracted section text recursively, excluded tabular content, and normalized section titles to canonical names prior to assembling SLM instructional prompts.
Extracted text was normalized across note schemas by removing control characters outside the printable ASCII range, standardizing whitespace, and reducing repeated line breaks while preserving paragraph boundaries.

For the DOM-like schemas, provided by the Oracle Health Millennium patient discharge summary template, the pipeline processed embedded nodes using a stack. 
When a schema node with the \texttt{ddsection} class is detected, the process extracts the section title from the first child node with class \texttt{ddsectiondisplay}. 
The child nodes were then processed recursively and accumulate text from nodes with \texttt{ddfreetext} or \texttt{blockformattedtext} classes, excluding any nodes identified as containing tabular elements.
The section titles were also normalized to map any variable representations with canonical section names to correct for variations in punctuation, spacing, and abbreviation. 
For example, "History of Present Illness", "HPI", and "history-of-present-illness" were mapped to the canonical identifier \texttt{history\_of\_present\_illness}.

We evaluated 26 open-source models spanning Llama, Qwen, Mistral, Gemma, Phi, and related variants, selected to vary domain adaptation, parameter size, and context length. Model identifiers and context windows are provided in Table~\ref{tab:ADD_llm_song_models_evaluated}.
Because the same prompt and scoring pipeline could be applied across model families, a broad set was selected to characterize model family and domain-adaptation effects.
Inference used deterministic decoding, model-specific loading APIs, post hoc JSON repair, and standardized mapping of model outputs to the predefined label matrix.

\begin{table}[htb!]
    \centering
    \fontsize{9pt}{9pt}\selectfont
    \setlength{\tabcolsep}{8pt}
    \caption{Evaluated models and context window lengths. }
    \vspace{-0.5em}
    \label{tab:ADD_llm_song_models_evaluated}
    \begin{threeparttable}
    \begin{tabularx}{\textwidth}{l r r l}
        \toprule
        \textbf{Model} & \textbf{Parameters} & \textbf{Context} & \textbf{Description} \\
        \midrule
        AdaptLLM/biomed-Qwen2.5-VL-3B-Instruct & 3B  & 32k   & Inherits Qwen2.5-VL 3B. \\
        deepseek-ai/DeepSeek-R1-Distill-Llama-8B & 8B & 128k & Llama 3.1–based distill. \\
        google/medgemma-4b-it                  & 4B  & 128k  & Google Med-Gemma instruct. \\
        HuggingFaceTB/SmolLM3-3B               & 3B  & 128k  & SmolLM3 long context. \\
        Joker-sxj/Qwen2.5-3B-instruct-medical-finetuned & 3B & 32k & Qwen2.5-3B-Instruct. \\
        meta-llama/Llama-3.1-8B-Instruct       & 8B  & 128k  & Llama 3.1 long context. \\
        meta-llama/Meta-Llama-3-8B-Instruct    & 8B  & 8k    & Llama 3 initial release. \\
        microsoft/Phi-4-mini-instruct          & mini & 128k & Phi-4 mini instruct. \\
        mistralai/Mistral-7B-Instruct-v0.2     & 7B  & 32k   & v0.2 context window. \\
        mosaicml/mpt-7b-instruct               & 7B  & 2k    & Short-form instruct. \\
        Qwen/Qwen2.5-7B-Instruct               & 7B  & 131k  & Full 128k+ context per card. \\
        Qwen/Qwen3-4B                          & 4B  & 32k   & Qwen3 base. \\
        Qwen/Qwen3-4B-Instruct-2507            & 4B  & 32k   & Qwen3 instruct. \\
        Qwen/Qwen3-8B-Base                     & 8B  & 128k  & Qwen3 8B base. \\
        TsinghuaC3I/Llama-3.1-8B-UltraMedical  & 8B  & 128k  & Llama 3.1 medical. \\
        unsloth/medgemma-4b-it-bnb-4bit        & 4B  & 128k  & Quant. Medgemma 4B IT. \\
        zai-org/glm-4-9b-chat-hf               & 9B  & 128k  & GLM-4-9B chat. \\
        lmsys/longchat-13b-16k                 & 13B & 16k   & LongChat 13B. \\
        Salesforce/xLAM-7b-r                   & 7B  & 32k   & Large Action Agent Model. \\
        MadeAgents/Hammer2.1-7b                & 7B  & 131k  & Qwen2.5-Coder-7B-Instruct. \\
        ibm-granite/granite-3.1-8b-instruct    & 8B  & 128k  & Granite 3.1 8B instruct. \\
        NousResearch/Hermes-3-Llama-3.1-8B     & 8B  & 128k  & Hermes 3 on Llama 3.1. \\
        allenai/Llama-3.1-Tulu-3-8B            & 8B  & 128k  & Tulu 3 on Llama 3.1. \\
        princeton-nlp/Llama-3-8B-ProLong-512k-Instruct & 8B  & 512k & Long-context Llama 3. \\
        01-ai/Yi-1.5-9B-32K                    & 9B  & 32k   & Yi-1.5 9B 32K. \\
        meetkai/functionary-small-v3.2         & 8B  & 128k  & Llama 3.1 8B base. \\
        \bottomrule
    \end{tabularx}
    \end{threeparttable}
\end{table}

\subsection*{Cloud Infrastructure and Secure PHI Processing}

\begin{figure}[htb!]
\centering
\includegraphics[width=0.82\textwidth]{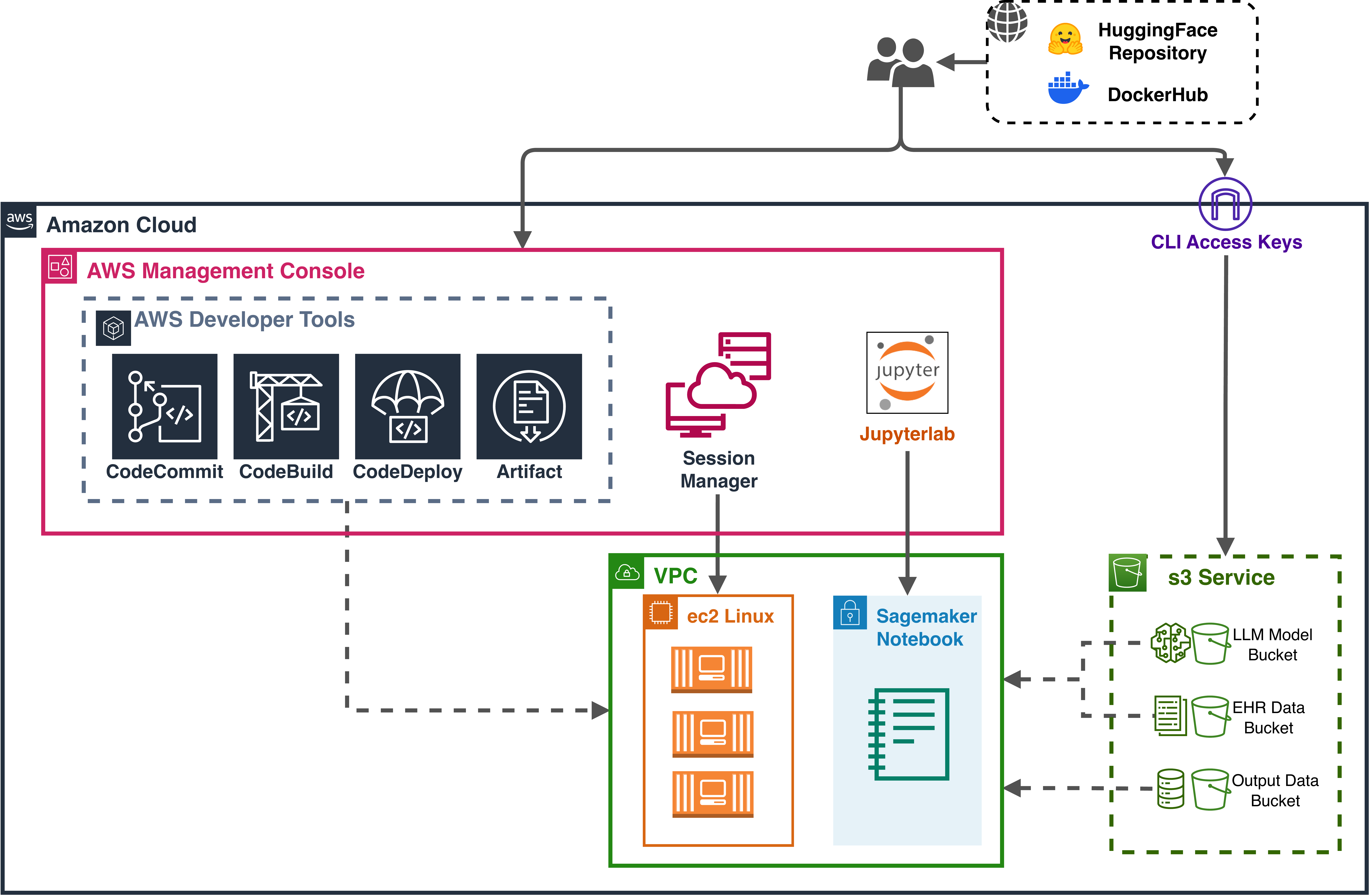}
\vspace{-1em}
\caption{AWS cloud infrastructure workflow for clinical information extraction.}
\label{fig:ADD_llm_song_revise_workflow}
\end{figure}

A computational instance for conducting SLM-based clinical information extraction was configured within a secure Amazon Web Services (AWS) SageMaker environment.
This was designed to maintain HIPAA compliance and data isolation for protected health information (PHI) within a workflow architecture, illustrated in Figure~\ref{fig:ADD_llm_song_revise_workflow}.
The infrastructure consisted of three primary layers for acquiring external models, module management, and isolated execution. 
External models were retrieved from HuggingFace model repositories as open-source SLM weights and their dependency files.
This separation was done so to isolate model artifacts before any loading and interaction with PHI occurs.

\subsection*{Dataset Description}

For evaluating the label detection pipeline, a development dataset was assembled using 200 note-section observations extracted from 23 patient discharge summaries across three ALS patients.
A closed 85-label ontology was selected for identifying key elements related to ALS care, which was then applied to SLM outputs, two simple automated non-generative baselines, and a manual human gold standard conducted in Label Studio Community Edition. 
The dataset contained 789 positive label observations, with a mean of 3.95 labels per observation, and 91 observations composed entirely of negative labels. 
These negative-only observations were intentionally retained because discharge-summary sections included tabular or otherwise non-narrative text without ALS-relevant target concepts, allowing validation to assess specificity and hallucination risk in addition to positive-label recall. 
Positive instances were concentrated in manual muscle test scores (216), medications (192), supplements (151), onset symptoms (106), forced vital capacity (39), ALSFRS-R total score labels (29), and ALSFRS-R subscore labels (26). 

\subsection*{Prompting and Evaluation Baselines}

SLMs were prompted to return minified JSON containing only the detected predefined ontology items. 
A corpus-wide item catalog defined the label space used for both prompting and evaluation. 
Model outputs were parsed and normalized into a binary presence matrix aligned to catalog items, which allowed direct multilabel comparison against the baselines and human gold standard label sets. 
We evaluated models using both aggregate and exact-match style metrics in order to distinguish broad label coverage from correctness at the note-section level.

The regex-derived baseline was constructed by applying case-insensitive patterns for canonical labels, common abbreviations, synonyms, and category-specific qualifiers such as FVC percentage notation to each note-section observation. This baseline produced a binary label matrix using deterministic pattern matches and was intended to represent a transparent rule-based comparator for standardized documentation elements.
The TF-IDF label-similarity baseline was constructed as a simple closed-ontology lexical comparator. 
Note-section text and label strings were represented with unigram and bigram TF-IDF features, and cosine similarity was computed between each note section and each ontology label.
Label-specific thresholds were selected within five-fold cross-validation to optimize F1 on the training folds before applying those thresholds to held-out observations. 
This baseline was included to compare SLM prompting with a non-generative similarity method within the same fixed label space. 

SLM outputs were processed through a multi-stage parsing and repair routine that normalized malformed keys, quotation marks, trailing commas, and non-JSON wrapper text before mapping outputs to catalog labels. 
All SLM outputs, the regex-derived rule baseline, and the TF-IDF label-similarity baseline were scored against the human gold standard label set as validation using micro-F1, macro-F1, precision, recall, exact-match accuracy, Hamming loss, and observation-level Jaccard similarity. 
Confidence intervals were estimated by observation-level bootstrapping.

\section*{Results}

\subsection*{Manual Validation and Overall Detection Performance}

\begin{figure}[htb!]
    \centerline{\includegraphics[width=\textwidth,height=0.85\textheight,keepaspectratio]{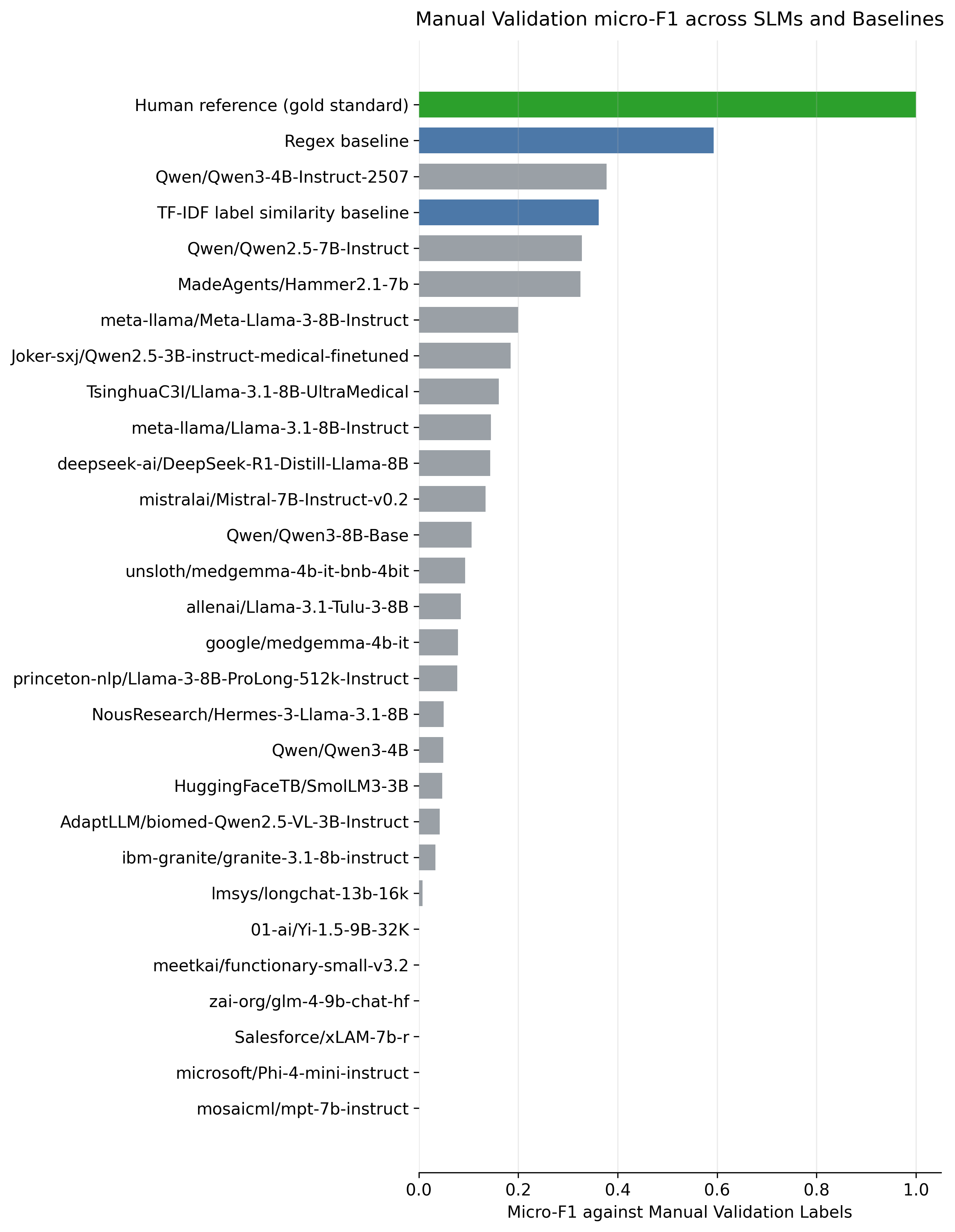}}
    \vspace{-1em}
    \caption{Manual-validation micro-F1 for the human reference, baselines, and all scored SLMs. The human reference is shown as a gold-standard performance ceiling rather than an extraction method.}
    \label{fig:manual_validation_micro_f1_ranking}
\end{figure}

Manual validation showed that SLMs were complementary to, but did not outperform, the baseline methods overall. 
As shown in Table~\ref{tab:manual_validation_all_models} and Figure~\ref{fig:manual_validation_micro_f1_ranking}, the regex baseline achieved the highest micro-F1 (0.593) and lowest Hamming loss (0.035) among detection methods. 
This suggests that several target concepts were sufficiently standardized for rule-based detection. 
The TF-IDF label-similarity baseline had substantially higher recall (0.843) but much lower precision (0.230), resulting in a micro-F1 of 0.362, exact-match accuracy of 0.0, and Hamming loss of 0.138. Among SLMs, Qwen3-4B-Instruct-2507 performed best by micro-F1 (0.378), while Qwen2.5-7B-Instruct and Hammer2.1-7B showed higher precision but lower recall.

\begin{table}[htb!]
\centering
\fontsize{9pt}{8pt}\selectfont
\setlength{\tabcolsep}{2pt}
\caption{Manual validation performance for the human reference, baselines, and all scored SLMs. Models are sorted by micro-F1.}
\vspace{-0.5em}
\label{tab:manual_validation_all_models}
\begin{threeparttable}
\begin{tabularx}{\textwidth}{X r r r r r r r r r r}
\toprule
\textbf{Model} & \textbf{TP} & \textbf{FP} & \textbf{TN} & \textbf{FN} & \textbf{Prec.} & \textbf{Rec.} & \textbf{Micro-F1} & \textbf{Macro-F1} & \textbf{Exact} & \textbf{Ham.} \\
\midrule
Human reference (gold standard) & 789 & 0 & 16211 & 0 & 1.0 & 1.0 & 1.0 & 1.0 & 1.0 & 0.0 \\
Regex baseline & 429 & 229 & 15982 & 360 & 0.652 & 0.544 & 0.593 & 0.468 & 0.475 & 0.035 \\
Qwen/Qwen3-4B-Instruct-2507 & 261 & 331 & 15880 & 528 & 0.441 & 0.331 & 0.378 & 0.295 & 0.440 & 0.051 \\
TF-IDF label similarity baseline & 665 & 2221 & 13990 & 124 & 0.230 & 0.843 & 0.362 & 0.588 & 0.0 & 0.138 \\
Qwen/Qwen2.5-7B-Instruct & 172 & 89 & 16122 & 617 & 0.659 & 0.218 & 0.328 & 0.160 & 0.455 & 0.042 \\
MadeAgents/Hammer2.1-7b & 157 & 19 & 16192 & 632 & 0.892 & 0.199 & 0.325 & 0.356 & 0.460 & 0.038 \\
meta-llama/Meta-Llama-3-8B-Instruct & 252 & 1479 & 14732 & 537 & 0.146 & 0.319 & 0.200 & 0.161 & 0.285 & 0.119 \\
Joker-sxj/Qwen2.5-3B-instruct-medical-finetuned & 164 & 826 & 15385 & 625 & 0.166 & 0.208 & 0.184 & 0.164 & 0.425 & 0.085 \\
TsinghuaC3I/Llama-3.1-8B-UltraMedical & 80 & 126 & 16085 & 709 & 0.388 & 0.101 & 0.161 & 0.067 & 0.455 & 0.049 \\
meta-llama/Llama-3.1-8B-Instruct & 171 & 1400 & 14811 & 618 & 0.109 & 0.217 & 0.145 & 0.100 & 0.315 & 0.119 \\
deepseek-ai/DeepSeek-R1-Distill-Llama-8B & 72 & 142 & 16069 & 717 & 0.336 & 0.091 & 0.144 & 0.068 & 0.410 & 0.051 \\
mistralai/Mistral-7B-Instruct-v0.2 & 109 & 727 & 15484 & 680 & 0.130 & 0.138 & 0.134 & 0.084 & 0.255 & 0.083 \\
Qwen/Qwen3-8B-Base & 92 & 853 & 15358 & 697 & 0.097 & 0.117 & 0.106 & 0.074 & 0.425 & 0.091 \\
unsloth/medgemma-4b-it-bnb-4bit & 84 & 935 & 15276 & 705 & 0.082 & 0.106 & 0.093 & 0.056 & 0.415 & 0.096 \\
allenai/Llama-3.1-Tulu-3-8B & 50 & 340 & 15871 & 739 & 0.128 & 0.063 & 0.085 & 0.047 & 0.290 & 0.063 \\
google/medgemma-4b-it & 78 & 1114 & 15097 & 711 & 0.065 & 0.099 & 0.079 & 0.049 & 0.395 & 0.107 \\
princeton-nlp/Llama-3-8B-ProLong-512k-Instruct & 60 & 699 & 15512 & 729 & 0.079 & 0.076 & 0.078 & 0.049 & 0.435 & 0.084 \\
NousResearch/Hermes-3-Llama-3.1-8B & 27 & 260 & 15951 & 762 & 0.094 & 0.034 & 0.050 & 0.019 & 0.410 & 0.060 \\
Qwen/Qwen3-4B & 20 & 5 & 16206 & 769 & 0.800 & 0.025 & 0.049 & 0.031 & 0.450 & 0.046 \\
HuggingFaceTB/SmolLM3-3B & 21 & 77 & 16134 & 768 & 0.214 & 0.027 & 0.047 & 0.044 & 0.455 & 0.050 \\
AdaptLLM/biomed-Qwen2.5-VL-3B-Instruct & 18 & 43 & 16168 & 771 & 0.295 & 0.023 & 0.042 & 0.019 & 0.455 & 0.048 \\
ibm-granite/granite-3.1-8b-instruct & 15 & 94 & 16117 & 774 & 0.138 & 0.019 & 0.033 & 0.016 & 0.450 & 0.051 \\
lmsys/longchat-13b-16k & 3 & 23 & 16188 & 786 & 0.115 & 0.004 & 0.007 & 0.002 & 0.450 & 0.048 \\
01-ai/Yi-1.5-9B-32K & 0 & 0 & 16211 & 789 & 0.0 & 0.0 & 0.0 & 0.0 & 0.455 & 0.046 \\
Salesforce/xLAM-7b-r & 0 & 0 & 16211 & 789 & 0.0 & 0.0 & 0.0 & 0.0 & 0.455 & 0.046 \\
microsoft/Phi-4-mini-instruct & 0 & 0 & 16211 & 789 & 0.0 & 0.0 & 0.0 & 0.0 & 0.455 & 0.046 \\
meetkai/functionary-small-v3.2 & 0 & 14 & 16197 & 789 & 0.0 & 0.0 & 0.0 & 0.0 & 0.450 & 0.047 \\
zai-org/glm-4-9b-chat-hf & 0 & 0 & 16211 & 789 & 0.0 & 0.0 & 0.0 & 0.0 & 0.455 & 0.046 \\
mosaicml/mpt-7b-instruct & 0 & 0 & 16211 & 789 & 0.0 & 0.0 & 0.0 & 0.0 & 0.455 & 0.046 \\
\bottomrule
\end{tabularx}
\begin{tablenotes}[flushleft]
\footnotesize
\item Abbreviations: TP, true positives; FP, false positives; TN, true negatives; FN, false negatives; Prec., precision; Rec., recall; Exact, exact-match accuracy; Ham., Hamming loss. Lower Hamming loss is better. The human reference row is a gold-standard self-comparison and represents the performance ceiling, not an extraction baseline. Exact-match accuracy should be interpreted alongside recall because negative-only observations can produce high exact-match values even when positive-label recall is low.
\end{tablenotes}
\end{threeparttable}
\end{table}

\subsection*{Precision--Recall Tradeoffs and Metric Sensitivity}

\begin{figure}[htb!]
    \centerline{\includegraphics[width=0.82\textwidth]{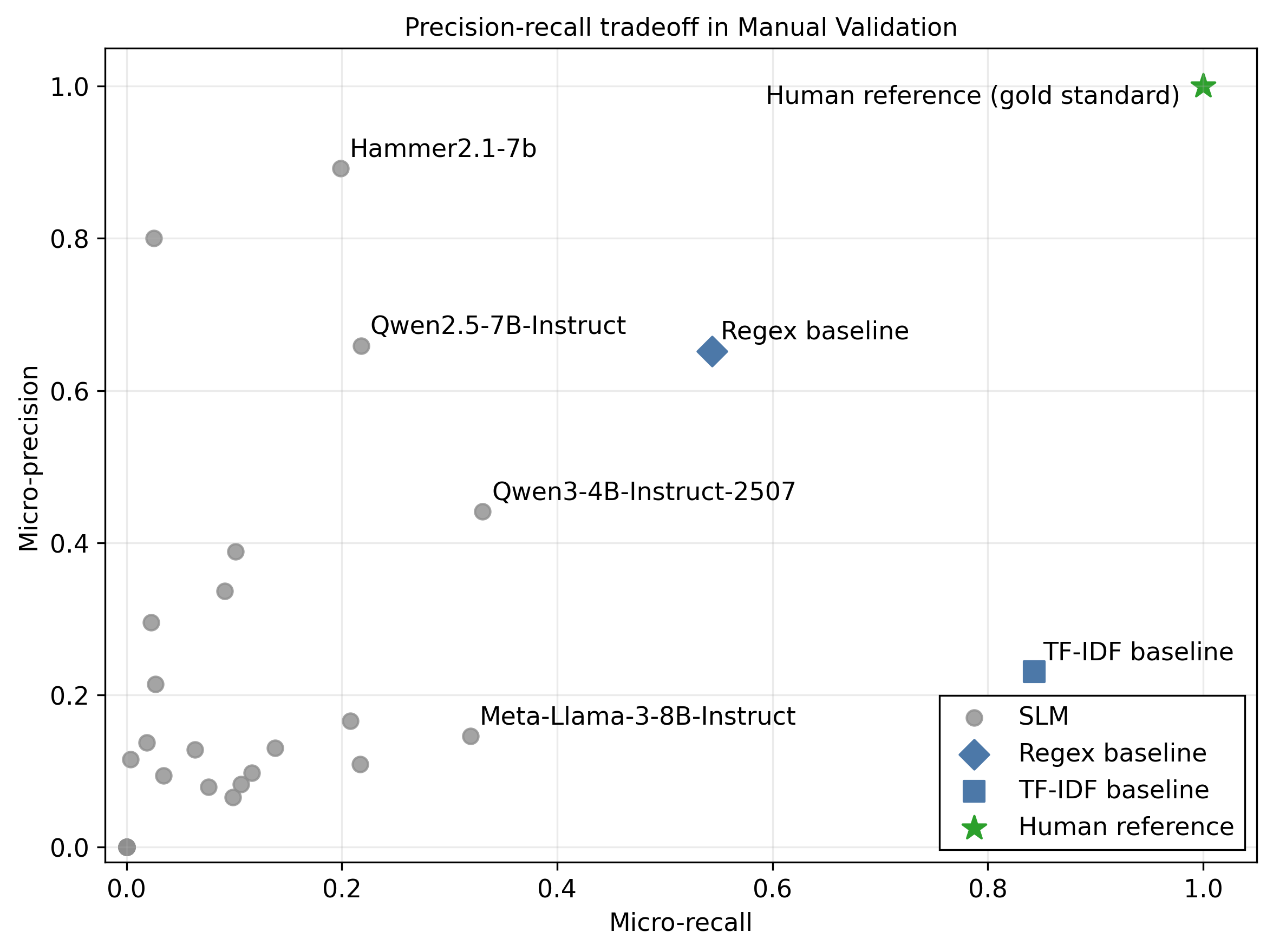}}
    \vspace{-1em}
    \caption{Precision-recall tradeoff across manually validated SLM outputs and baseline comparators.}
    \label{fig:precision_recall_tradeoff}
\end{figure}

Figure~\ref{fig:precision_recall_tradeoff} shows that the main difference between methods was the balance between missed labels and false-positive labels. The regex baseline had the most balanced precision-recall profile among the automated baselines, with precision of 0.652 and recall of 0.544. In contrast, TF-IDF recovered more true labels but produced many false positives, reflecting a broad but less selective labeling strategy. Qwen3-4B-Instruct-2507 was more precise than TF-IDF but missed more positive labels, while Hammer2.1-7B and Qwen2.5-7B-Instruct were more conservative and produced fewer false positives. These differences are clinically relevant because high-recall approaches may be useful for broad screening, whereas high-precision approaches may be preferable when the goal is to reduce manual review burden.

\begin{figure}[htb!]
    \centerline{\includegraphics[width=\textwidth,height=0.85\textheight,keepaspectratio]{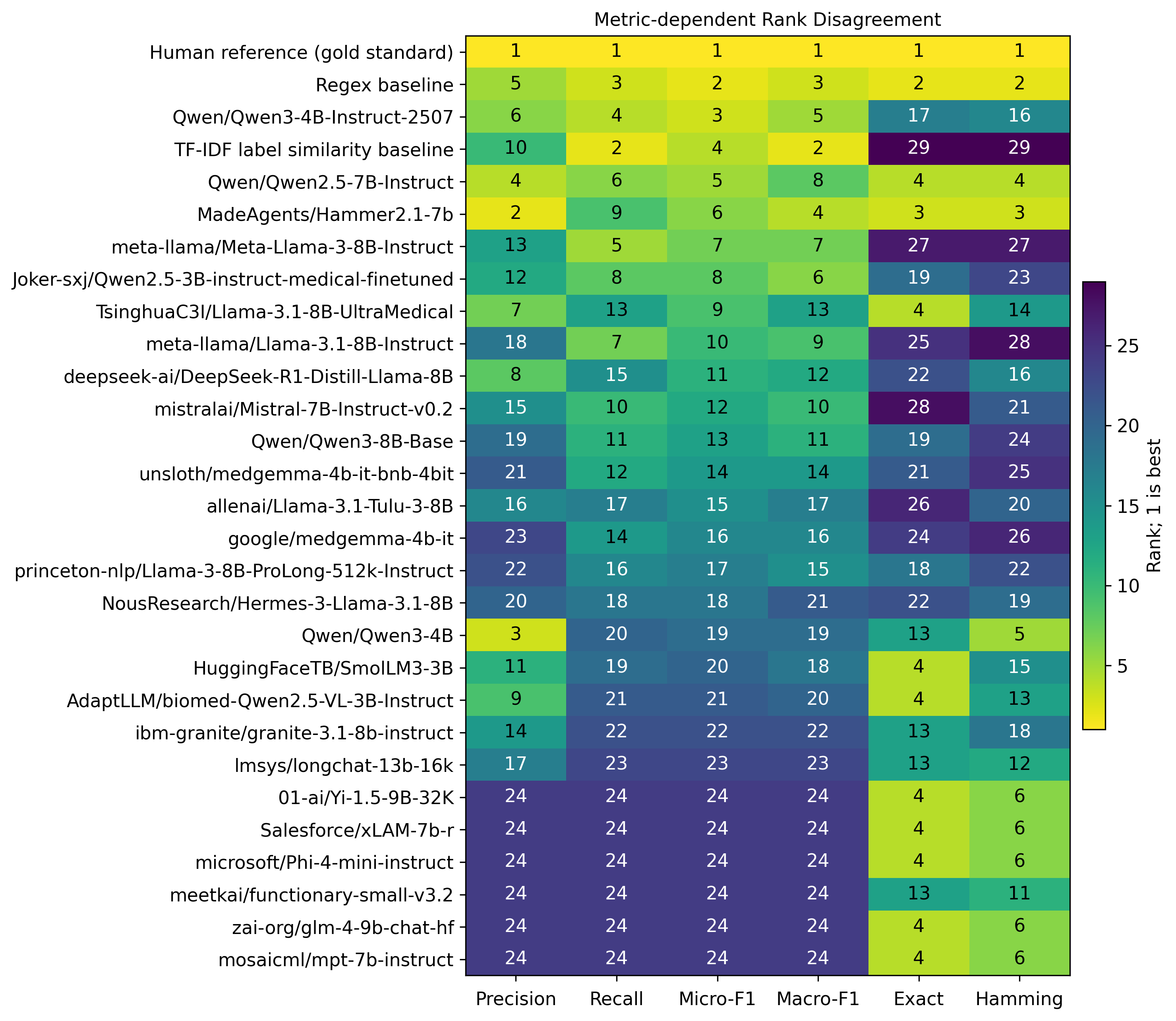}}
    \vspace{-1em}
    \caption{Model rankings differed by validation metric, showing that precision, recall, micro-F1, macro-F1, exact-match accuracy, and Hamming loss emphasize different error profiles.}
    \label{fig:metric_rank_disagreement}
\end{figure}

Metric-dependent rank disagreement further showed that no single score fully captured model behavior, as shown in Figure~\ref{fig:metric_rank_disagreement}. The TF-IDF baseline ranked highly by recall and macro-F1, but poorly by exact-match accuracy and Hamming loss because extensive over-labeling produced the highest false-positive rate among all methods. Conversely, models with stronger precision or lower Hamming loss did not necessarily rank highest by recall or micro-F1. Therefore, micro-F1 should be interpreted as a useful overall summary, but model selection should also consider the intended use case and the relative costs of false positives, missed labels, and incomplete section-level predictions.

\subsection*{Category-Level Validation}

\begin{figure}[htb!]
    \centerline{\includegraphics[width=\textwidth,height=0.45\textheight,keepaspectratio]{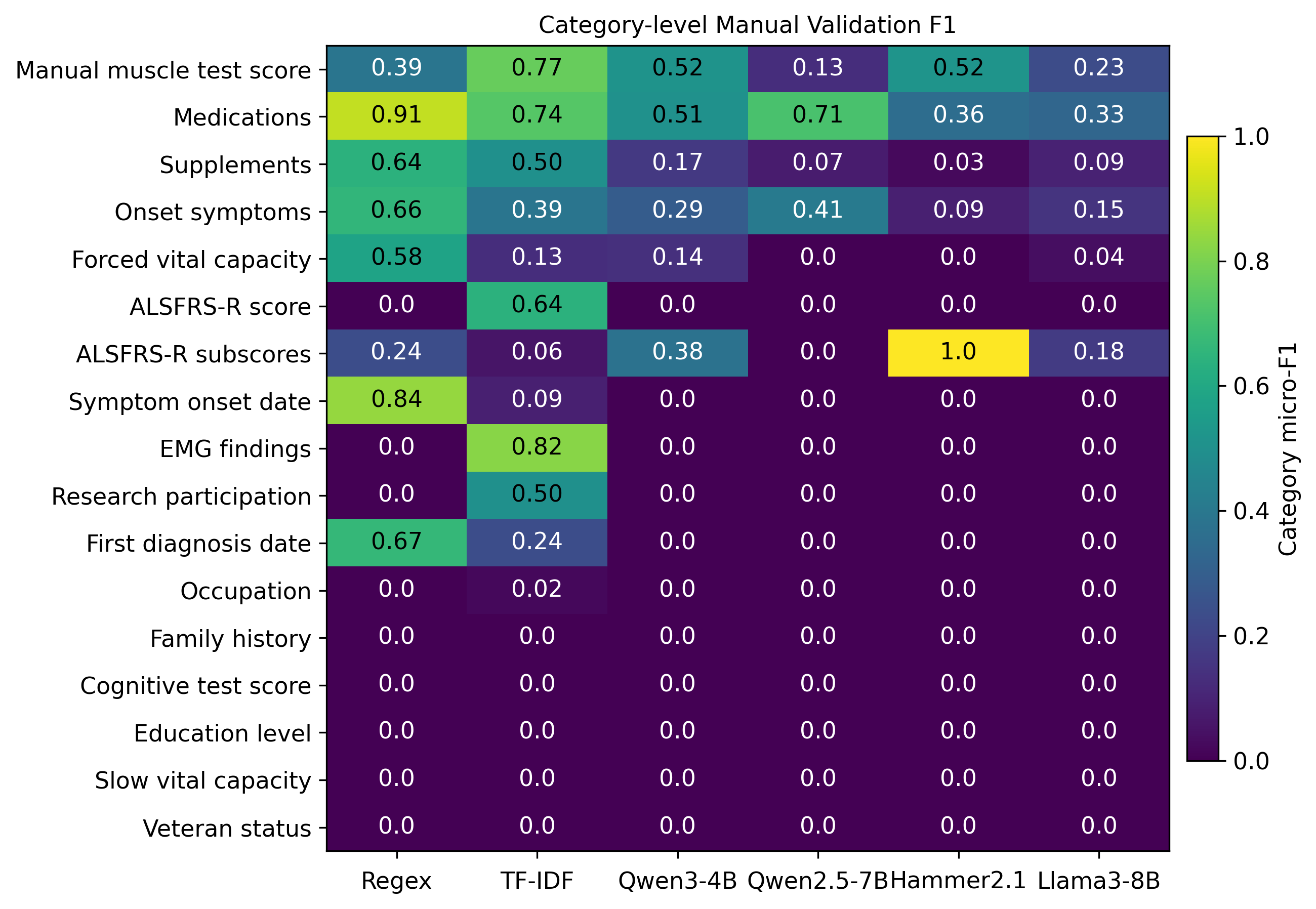}}
    \vspace{-1em}
    \caption{Category-level manual-validation F1 for baselines and selected high-performing SLMs.}
    \label{fig:category_f1_key_configurations}
\end{figure}

Category-level validation showed that performance depended primarily on the type of label being identified, as shown in Figure~\ref{fig:category_f1_key_configurations}. No method was best across all categories. Hammer2.1-7B performed best for ALSFRS-R subscores, whereas the regex baseline performed best for several categories with more predictable phrasing, including medications, supplements, onset symptoms, forced vital capacity, symptom onset date, and first diagnosis date. The TF-IDF baseline performed well for categories where note text closely matched the label, such as manual muscle test score, ALSFRS-R score, EMG findings, and research participation, but this came at the cost of high false positives. Several categories remained difficult for all methods, especially sparse labels, numeric-heavy fields, respiratory measures, manual muscle testing, and statements involving negation or absence. These results suggest that the most practical approach may be to match the detection method to the label type rather than selecting one model for all labels.
Regex-based rules may be used for standardized terms, lexical similarity for broad screening, and SLMs for categories with higher phrase variation.

\section*{Discussion}

\subsection*{Clinical Implementation Considerations}

The primary clinical implication is that model choice depends on the structure of the targeted narrative text and acceptable error. 
The manual validation results did not support replacing rule-based or simple embedding methods with any one SLM across all labels. 
Instead, they suggest that rule-based detection remains useful for standardized documentation patterns, TF-IDF label similarity can serve as a high-recall but low-precision comparator, and SLMs may be most valuable for selected categories where phrasing is variable such as in semi-structured text. 
This distinction is important because the task is a closed-ontology multilabel classification problem rather than open-ended clinical concept discovery. 
All evaluated approaches map heterogeneous text into a predefined label ontology space. 

\paragraph{Model selection strategies}

For broad first-stage term-presence screening, the regex baseline might remain an important comparator because it performed best overall in manual validation. 
TF-IDF label similarity may be useful when recall is prioritized and downstream review can absorb many false positives, but its exact-match accuracy and Hamming loss indicate that it should not be used alone for observation-level detection. 
Qwen3-4B-Instruct-2507 was the best single SLM by micro-F1, but its lower performance compared with the regex baseline indicates that prompt-only SLM detection needs to be validated before deployment. 
Qwen2.5-7B-Instruct and Hammer2.1-7B achieved higher precision but lower recall, suggesting they may be better suited for scenarios where minimizing false positives is especially critical.
Hammer2.1-7b was best at ALSFRS-R subscore detection, supporting targeted model use for selected functional assessment labels instead of as a general-purpose replacement strategy. 
For applications that will not tune model choice extensively, a practical workflow would be to begin with a rule-based baseline for standardized labels, compare it with one high-recall lexical baseline and one strong general-purpose SLM on a small local validation set, and then add category-specific SLM extraction where it improves the error required by the intended workflow. \\

These results also underscore the importance of using multiple reporting metrics over a single summary score.
Micro-F1 summarizes frequency-weighted label performance, macro-F1 gives more influence to sparse labels, exact-match accuracy evaluates full-observation correctness, and Hamming loss captures the average label-level error rate. 
Because these metrics measure different aspects of behavior, model choice should depend on the intended use. 
A screening workflow may favor recall, while a workflow meant to reduce manual review may favor precision or low Hamming loss. 
A workflow that requires all labels for a note section to be correct should place more weight on exact-match accuracy. 
The different strengths of the baselines and SLMs suggest that ensemble or routing approaches may be useful, but we did not test such a system here. 
For that reason, we treat ensembles as future work rather than as a recommendation from this study.

\subsection*{Limitations}

The method described has some important limitations. 
SLM outputs are probabilistic and can include hallucinations, in which the model reports items that are absent from the source text. 
Although deterministic decoding was used to reduce output variability, this does not eliminate false-positive or false-negative errors. 
The regex and TF-IDF baselines also have important limitations. 
Regex patterns can miss nonstandard phrasing, while TF-IDF label similarity can over-label sections when lexical overlap is high but the target concept is absent or contextually different, as was demonstrated in this study.
Manual validation was limited to the small number of note-section observations available, so the validation results should be interpreted as evidence about comparative behavior and error patterns rather than as determinative of model accuracy.

JSON parsing and correction introduces additional variability and extraction completeness. 
Despite using a multi-stage repair pipeline, some model outputs were unparseable. 
The minimal fallback strategy (returning empty elements) prevents complete failures but induces information loss from malformed outputs. 
More analysis is needed to better understand which models produce well-formed outputs from those requiring correction, which could serve as an indicator of model reliability.
The item catalog discovery process relied on a corpus-wide ruleset that could overlook sparse items or include irrelevant items that appear more frequently. 
The current task also evaluated term presence rather than value-level correctness, so labels such as forced vital capacity, manual muscle testing grades, and ALSFRS-R scores require additional validation before being used as structured clinical measurements.

\subsection*{Future Directions}

Including value extraction prompts after presence detection would create a two-stage pipeline that first identifies terms and then extracts their values when present. 
This hierarchical approach may improve efficiency by processing only sections found to contain relevant information, while avoiding additional computation or value searching in sections where terms are absent. 
An active learning strategy could also be implemented, with model selection performed iteratively and potentially enhanced through a human-in-the-loop approach. 
Models could first be evaluated on smaller pre-labeled subsets, after which the best-performing model could be selected for full-corpus extraction. 
Tabulated content could be processed using a specialized prompt template or transformed into a text representation, such as a dictionary or tuple object, that preserves row-column relationships. 
Table-specific models trained for structured data extraction may also be useful. Together, these steps could improve the completeness of information captured from note sections in patient discharge summaries. \\

\section*{Conclusion}
Open-source small language models can support first-stage extraction of ALS-relevant clinical term presence from heterogeneous narrative documentation without task-specific supervised training. However, manual validation showed that prompt-only SLM extraction did not uniformly outperform automated non-generative baseline techniques. The regex baseline performed best overall by micro-F1 and Hamming loss, whereas the TF-IDF label-similarity baseline showed high recall but substantial over-labeling. Performance differed by metric and clinical label type, with Qwen3-4B-Instruct-2507 performing best among single SLMs by micro-F1 and Hammer2.1-7b showing targeted use for ALSFRS-R subscore detection. These findings support hybrid extraction workflows in which rule-based methods are retained for standardized terms, lexical methods are interpreted as high-recall comparators, and SLMs are added selectively for categories with more variable documentation. Future work should extend validation to value-level extraction and larger manually annotated samples before clinical deployment.


\subparagraph{Acknowledgments}We acknowledge the use of the HABITAT (Health Analytics and Biomedical Informatics Trusted Access and Technology) platform at the Center for Biomedical Informatics, University of Missouri, for supporting the computational environment used to conduct these experiments. This work was supported through the Center of Disease and Control (CDC) (\#R01TS000336), the Department of Defense office of the Congressionally Directed Medical Research Programs (CDMRP) through the Amyotrophic Lateral Sclerosis Research Program (ALSRP) Clinical Development Award (\#W81XWH-22-1-0491), and the ALS Association (\#24-AT-722). Opinions, interpretations, conclusions, and recommendations are those of the authors and are not necessarily endorsed by the Department of Defense. 

\makeatletter
\renewcommand{\@biblabel}[1]{\hfill #1.}
\makeatother

\bibliographystyle{vancouver}
\bibliography{references}

\end{document}